\documentclass{article}

    \PassOptionsToPackage{numbers, sort&compress}{natbib}

\usepackage[preprint]{neurips_2024}

\usepackage[utf8]{inputenc} 
\usepackage[T1]{fontenc}    
\usepackage{hyperref}       
\usepackage{url}            
\usepackage{booktabs}       
\usepackage{amsfonts}       
\usepackage{nicefrac}       
\usepackage{microtype}      
\usepackage{xcolor}         
\usepackage{bbding}
\usepackage{graphicx}
\usepackage{amsmath}
\usepackage{algorithm}
\usepackage{algorithmicx}
\usepackage{algpseudocode}
\usepackage{bm}
\usepackage{multirow}
\usepackage{threeparttable}
\usepackage{makecell}
\usepackage{wrapfig}
\usepackage{subcaption}
\usepackage{array}
\usepackage{utfsym}
\usepackage[misc]{ifsym} 

\makeatletter
\newcommand\figcaption{\def\@captype{figure}\caption}
\newcommand\tabcaption{\def\@captype{table}\caption}
\newcommand{\tabincell}[2]{\begin{tabular}{@{}#1@{}}#2\end{tabular}}
\makeatother

\title{TabPedia: Towards Comprehensive Visual Table Understanding with Concept Synergy}

\author{
 \textnormal{
 \normalsize{Weichao Zhao$^{1,2,\spadesuit,}$\thanks{Equal contribution. $\spadesuit$ \text{Interns at ByteDance.} $\ddag$ \text{Project lead.}}
 ~~~Hao Feng$^{1,\ast}$
 ~~~Qi Liu$^{2,\ast}$
 ~~~Jingqun Tang$^{2}$
 ~~~Shu Wei$^{2}$
 ~~~Binghong Wu$^{2}$
 }
 }
\\
\normalsize{
Lei Liao$^{2}$
~~~Yongjie Ye$^{2}$
~~~Hao Liu$^{2,\ddag,}$\thanks{\textrm{\Letter}~Corresponding authors: Wengang Zhou and Hao Liu.}
~~~Wengang Zhou\textsuperscript{\rm 1,}$^\dag$ 
~~~Houqiang Li$^{1}$
~~~Can Huang\textsuperscript{\rm 2}}
\\
\textnormal{\small{$^{1}$ University of Science and Technology of China, $^{2}$ ByteDance Inc.,}}
\\
\small{\{saruka, haof\}@mail.ustc.edu.cn, \{zhwg, lihq\}@ustc.edu.cn}
\\
\small{\{liuqi.nero, haoliu.0128, can.huang\}@bytedance.com}
}

\begin{document}

\maketitle

\begin{abstract}

    Tables contain factual and quantitative data accompanied by various structures and contents that pose challenges for machine comprehension. 
    Previous methods generally design task-specific architectures and objectives for individual tasks, resulting in modal isolation and intricate workflows.
    In this paper, we present a novel large vision-language model, TabPedia, equipped with a \textit{concept synergy} mechanism. In this mechanism, all the involved diverse visual table understanding (VTU) tasks and multi-source visual embeddings are abstracted as concepts. This unified framework allows TabPedia to seamlessly integrate VTU tasks, such as table detection, table structure recognition, table querying, and table question answering, by leveraging the capabilities of large language models (LLMs). Moreover, the concept synergy mechanism enables table perception-related and comprehension-related tasks to work in harmony, as they can effectively leverage the needed clues from the corresponding source perception embeddings. Furthermore, to better evaluate the VTU task in real-world scenarios, we establish a new and comprehensive table VQA benchmark, ComTQA, featuring approximately 9,000 QA pairs. Extensive quantitative and qualitative experiments on both table perception and comprehension tasks, conducted across various public benchmarks, validate the effectiveness of our TabPedia. The superior performance further confirms the feasibility of using LLMs for understanding visual tables when all concepts work in synergy. The benchmark ComTQA has been open-sourced at \href{https://huggingface.co/datasets/ByteDance/ComTQA}{https://huggingface.co/datasets/ByteDance/ComTQA}. The source code and model also have been released at \href{https://github.com/zhaowc-ustc/TabPedia}{https://github.com/zhaowc-ustc/TabPedia}.

\end{abstract}

\section{Introduction}
With the rapid advancement of digital technology, numerous paper documents must be converted into electronic formats for efficient storage and utilization. Tables, as indispensable components of documents, play a vital role in summarizing facts and quantitative data \cite{Gao_ICDAR_2019, Gobel_ICDAR_2013}. The compact yet informative nature of tables makes them advantageous for various applications, thereby attracting widespread research attention toward Visual Table Understanding (VTU). VTU generally encompasses four subtasks: \textit{Table Detection} (TD), which locates tables within document images; \textit{Table Structure Recognition} (TSR), which parses the structure of tables in table-centric images; \textit{Table Querying} (TQ), which recognizes the structure of a table from an entire image at a given location, a task that remains underexplored in the previous works; and \textit{Table Question Answering} (TQA), which answers questions based on table contents. These tasks pose challenges from various perspectives due to the need for representations at different visual-semantic granularities and hierarchies.

\begin{figure}[t]
	\centering
	\includegraphics[width=1.0\textwidth]{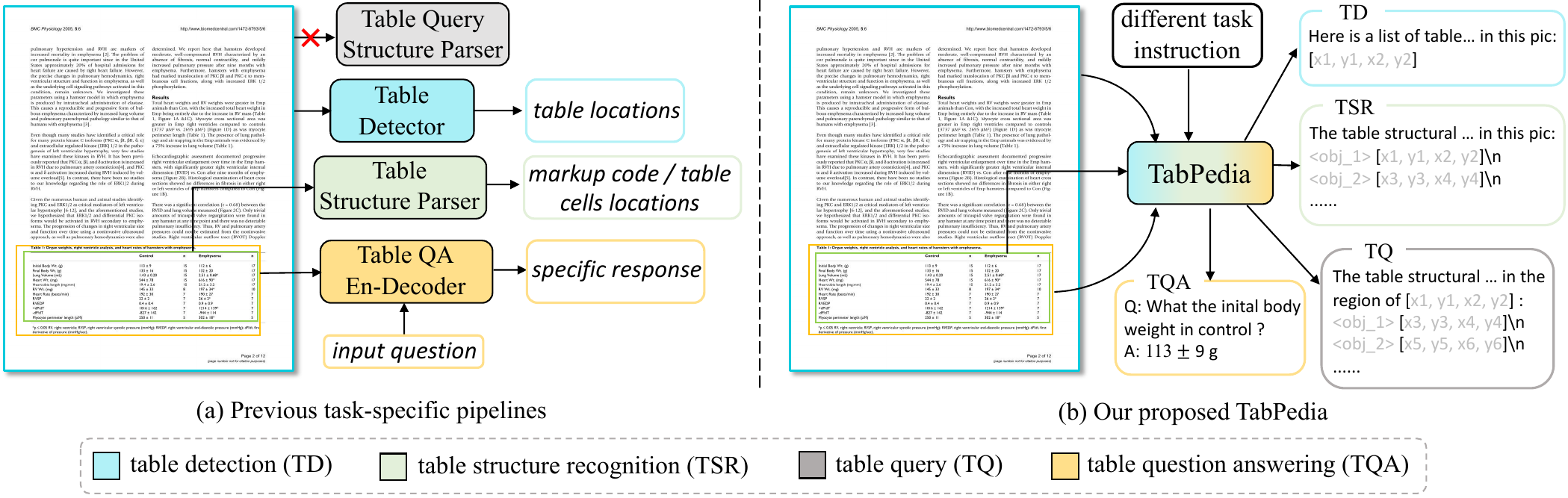} 
	\vspace{-1.5em}
	\caption{Comparison with previous task-specific pipelines for visual table understanding. In contrast to design different architectures for various table tasks, our TabPedia effectively performs these tasks in a unified framework through delicately leveraging the understanding capability of LLMs.}
	\vspace{-1.2em}
	\label{intro}
\end{figure}

Given the success achieved, many pioneering works have mainly centered on the specific subtask with various task-specific architectures, as shown in Fig.~\ref{intro} (a). For visual table perception tasks such as TD and TSR, one of most adopted approaches is in the detection manner~\cite{Prasad_CascadeTabNet_2020,schreiber2017deepdesrt,zheng2021global,siddiqui2018decnt,nguyen2022tablesegnet,zhang2023yolo,smock2022pubtables}. In contrast, generative vision-language models~\cite{feng2023docpedia,bai2023qwen,liu2024textmonkey,mPLUG-DocOwl} are often employed to generate answers conditioned on the semantic content of tables for TQA task. Specifically, Vision Transformers (ViT) \cite{dosovitskiy2020image} pretrained on CLIP \cite{radford2021learning} or EVA-CLIP \cite{fang2023eva}, Swin-Transformer \cite{liu2021swin}, and similar models serve as vision encoders, while language models operate in either encoder-decoder \cite{lewis2020bart, raffel2020exploring} or decoder-only frameworks \cite{radford2018improving, brown2020language, touvron2023llama, bai2023qwen}. Besides, recent fast-growing Large Vision Language Models~(LVLMs)~\cite{alayrac2022flamingo,chen2022pali, blip2,dai2023instructblip,kosmos,kosmos2,zhu2023minigpt,liu2023visual,ye2023mplug,mPLUG-DocOwl,shikra,li2023otter,yang2023dawn,bai2023qwen} have shown their powerful capabilities to perceive and understand visual clues by integrating instruction following of Large Language Models~(LLMs)~\cite{gpt4,gpt3,anil2023palm,gao2023llama,qwen7b}. Despite impressive progress, the \textit{status quo} begs for a question: ``\textit{Can we leverage the advantages of LVLMs to solve all the VTU tasks once and for all?}'' 

A straightforward solution would be to train the LVLM directly using all the VTU data. However, aside from the diverse table structure and the various relations of table contents, it remains a nontrivial issue due to two cruxes of table parsing and understanding: (i)~discrepancy between the representation formats ~(two-dimensional structure VS. one-dimensional sequence); (ii)~required image resolutions. Although some works~\cite{hu2024mplug,wan2024omniparser,peng2024unitable} represent table structure in markup formats like HTML, XML, Markdown, or LATEX. However, they neglect spatial coordinates for cells and only encode logical relationships implicitly. The generated code contains extensive formatted information from different markup languages, increasing output length and potentially causing parsing issues with illegal grammars.

To attack above issues, we in this paper propose a novel LVLM tailored for comprehensive VTU, TabPedia, to effectively solve all VTU tasks in a unified framework, as shown in Fig.~\ref{intro}~(b). More concretely, we employ dual vision encoders, namely ViT-L~\cite{radford2021learning} and Swin-B~\cite{kim2022ocr}, to encode the global and fine-grained local information in the low- and high-resolution formats of the input image respectively, acquiring multi-source visual embeddings. Here, all the involved VTU tasks and multi-source visual embeddings are abstracted as \textit{concepts} and \textit{concept synergy} mechanism is implemented by introducing the \textit{mediative tokens} to the LLM in our model. Thanks to this mechanism, all the concepts in TabPedia can work in synergy flexibly. Quantitative and qualitative experimental results on both table perception and comprehension tasks across various public benchmarks confirm the effectiveness of our proposed TabPedia. To further investigate the potential of our model in more challenging and realistic scenarios, we establish a new and comprehensive table VQA benchmark, ComTQA, featuring round 1,500 images and 9,000 QA pairs.

Our contributions are summarized as follows,
\begin{itemize}
    \item We propose a novel large vision-language model, TabPedia, to integrate various VTU tasks into a unified framework, including TD, TSR, TQ and TQA. Specifically, TabPedia fully leverages the comprehensive capabilities of LLMs to fertilize complex table understanding.

    \item We design a concept synergy mechanism to harmonize both table perception and comprehension tasks.
    Through introducing the meditative tokens into our framework, TabPedia adaptively enables useful information in multi-source visual embeddings and task instructions, generating accurate and plausible responses.

    \item Extensive quantitative and qualitative experiments validate the effectiveness of our proposed TabPedia across various tasks and benchmarks. To further exploit the potential of our model in more complex scenarios, we build a new table VQA benchmark, ComTQA, involving multiple answers, mathematical calculation and logical reasoning, \textit{etc}.
\end{itemize}

\section{Related Work}

\subsection{Table Recognition}
\vspace{-0.5em}
Table recognition is generally divided into table detection, table structure recognition and table content recognition
In our work, table content recognition is beyond our scope.

For TD task, the earliest approaches are rule-based methods for locating tables inside documents~\cite{kieninger1999t,gatos2005automatic,harit2012table}. 
With the rapid advances in deep learning, numerous CNN-based methods show impressive performance. 
Most of these methods directly adopt top-down object detection frameworks to solve this problem~\cite{vo2018ensemble,gilani2017table,huang2019yolo,zheng2021global,prasad2020cascadetabnet,agarwal2021cdec,sun2019faster}.
For instance, Sun~\textit{et al.}~\cite{sun2019faster} adopt Faster R-CNN~\cite{sun2019faster} to detect table boxes and the
corresponding corner boxes simultaneously, and then adjust table boundaries according to the detected corners.
Some other methods model each document image as a graph and formulate TD as a graph labeling problem~\cite{riba2019table,holevcek2019table,li2020docbank}.
In addition, TATR~\cite{smock2022pubtables} first applies the transformer-based detector, DETR~\cite{carion2020end}, to improve the detection accuracy without special customization. 

For TSR task, one of the most common modeling approaches is still to regard it as some form of object detection~\cite{Prasad_CascadeTabNet_2020,schreiber2017deepdesrt,zheng2021global,smock2022pubtables,guo2022trust,lin2022tsrformer,liu2024grab}. Among them, DeepDeSRT~\cite{schreiber2017deepdesrt} and TableNet~\cite{paliwal2019tablenet} are both representative works exploring semantic segmentation to obtain table cell boundaries.
TATR~\cite{smock2022pubtables} first proposes to utilize DETR for this task. 
TSRFormer~\cite{lin2022tsrformer} introduces a cross-attention module into the DETR framework to improve the localization accuracy of row/column separators.
Some other methods attempt to parse table structure via modeling relationship among different table elements~\cite{Liu2021show,chi2019complicated,liu2022neural}.
As the most relevant to our approach, markup generation-based methods directly generate markup (HTML or LaTeX) sequences from raw table images~\cite {zhong2020image,wan2024omniparser}. EDD~\cite{zhong2020image} introduces a cell decoder and a structures decoder to generate HTML codes. OmniParser~\cite{wan2024omniparser} further integrates three task-specific decoders to enhance the table structure representation.

While the previous methods have achieved promising results on table perceptive tasks, they are still limited in table intricate content understanding.
In our work, we jointly exploit table perception and comprehension tasks in a unified framework, concurrently enriching visual table understanding.

\vspace{-0.5em}
\subsection{Large Vision-Language Models}
\vspace{-0.5em}
LVLMs aim to equip LLMs~\cite{chiang2023vicuna,gao2023llama,zhu2023minigpt,gpt3,qwen7b} with visual comprehension capability. 
The mainstream approaches attempt to connect visual encoders and LLMs with intermediate modules such as simple Projectors~\cite{liu2023visual}, QFormer~\cite{blip2},  Perceiver Resamplers~\cite{alayrac2022flamingo}, achieving visual language understanding through pre-training alignment and instruction fine-tuning.
For text-rich document scene, several works~\cite{mPLUG-DocOwl,feng2023docpedia,feng2023unidoc,wei2023vary,hu2024mplug,wan2024omniparser,fujitake2024layoutllm} propose to enhance the LVLMs' capabilities in understanding textual elements~(text-centric VQA, OCR, text spotting, \textit{etc.}).
Among them, TextMonkey~\cite{liu2024textmonkey} employs shifted window attention and token resampler module to improve the training process. 
DocOwl-1.5~\cite{hu2024mplug} collects a comprehensive dataset DocStruct4M to support unified structure learning.

Despite achieving extraordinary progress on visual understanding, existing LVLMs still face challenges in two-dimensional table parsing and understanding. In this paper, we propose a unified framework to concurrently achieve table perception and comprehension with the support of LLMs.

\vspace{-0.5em}
\subsection{Additional Tokens}
\vspace{-0.5em}
In the trend of Transformer-based approaches, extending the input sequence with special tokens is popularized for various intentions, such as extracting task-specific information~\cite{dosovitskiy2020image,carion2020end}, providing extra information~\cite{ge2023context,xue2023adaptive} or improving model performance~\cite{burtsev2020memory,bulatov2022recurrent,darcet2023vision,goyal2023think}.
For instance, ViT~\cite{dosovitskiy2020image} utilizes [CLS] token for classification.
Similarly, DETR~\cite{carion2020end} proposes object queries for detection.
ATR~\cite{xue2023adaptive} adopts tape tokens to obtain useful information from a memory bank.
In addition, the Memory Transformer~\cite{burtsev2020memory} presents a simple approach to improve translation performance by attaching trainable memory tokens after the token sequence. Darcet~\textit{et al.,}~\cite{darcet2023vision} further attempt to add extra tokens in ViT-based frameworks, \textit{e.g.,} CLIP~\cite{radford2021learning} and DINOv2~\cite{oquab2023dinov2}, thus improving visual tasks.
In our work, we inherit this spirit and design meditative tokens to enhance TabPedia's perceptive and comprehensive capability for visual tables.

\begin{figure}[t]
	\centering
	\includegraphics[width=1.0\textwidth]{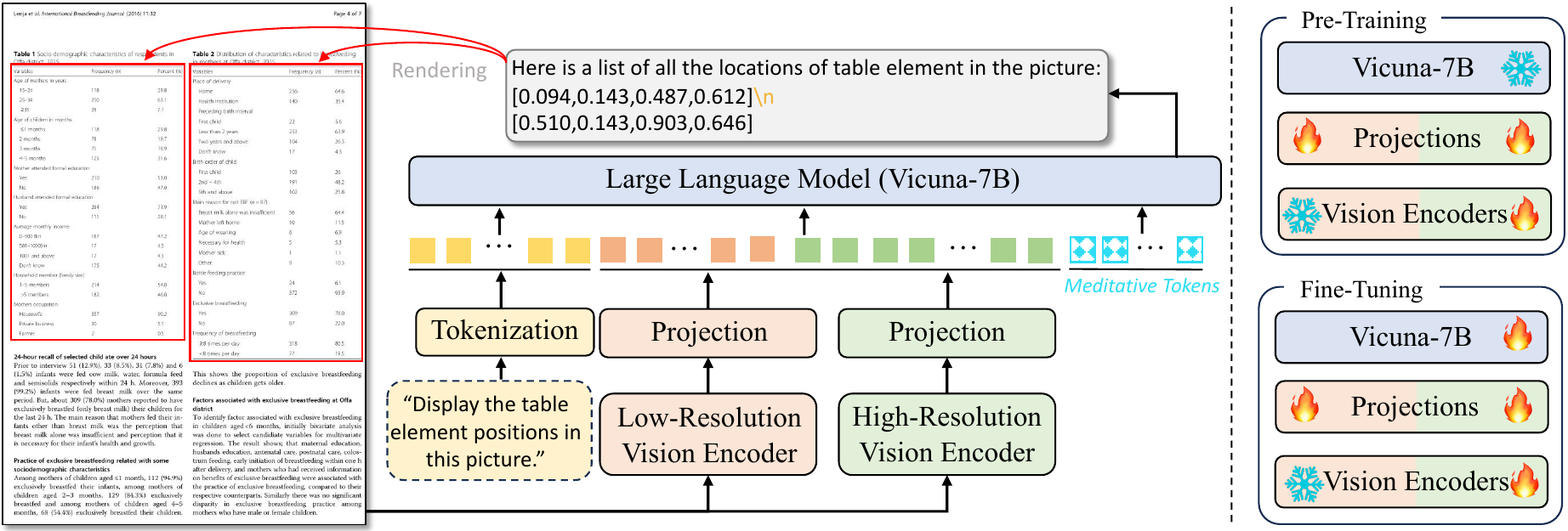} 
	\vspace{-0.2em}
	\caption{The illustration of our proposed TabPedia. Given the input image, TabPedia feeds it into both vision encoders attached projections to extract different granular features. Then, the visual tokens are combined with instruction-derived tokens, and fed into the LLM. The LLM leverages its powerful understanding ability to generate a plausible response.}
	\vspace{-1.0em}
	\label{framework}
\end{figure}

\section{Method}
As shown in Fig~\ref{framework}, we present an overview of TabPedia. The overall training pipeline consists of two phases. Concretely, the pre-training stage aims to align the visual features to the large language model, and the fine-tuning stage focuses on visual table-aware understanding. In the following, we elaborate on the architecture of TabPedia, followed by the exposition of its two training phases.

\subsection{Model Architecture}
\textbf{High-Resolution Vision Encoder.} As proved by previous methods~\cite{kim2022ocr,lee2023pix2struct,ye2023ureader}, the high-resolution image is critical to ensuring that the LLMs could grasp rich visual information. Following Donut~\cite{kim2022ocr}, we adopt Swin-B~\cite{liu2021swin} to encode the high-resolution format of input image. 
Given the input RGB image $\mathit{I}$, we first resize it to pre-defined high-resolution scale of $\mathrm{H}\times \mathrm{W}$, denoted as $\mathit{I}_{h}$. 
By default, both $\mathrm{H}$ and $\mathrm{W}$ are set to 2,560 and 1,920, respectively. 
Notably, we maintain the aspect ratio during the resizing process to prevent distortion of table contents and structures.
Then, the resized image $\mathit{I}_{h}$ is fed into the vanilla Swin Transformer initialized from~\cite{kim2022ocr} to obtain a feature map $\mathit{V}_h$ downsampled by a factor of 1/32, each token with 1,024 dimension.

\textbf{Low-Resolution Vision Encoder.} To keep the overall layout information, the raw image is also resized to a low-resolution one denoted as $\mathit{I}_{l}$. We choose the pre-trained CLIP visual encoder ViT-L/14~\cite{radford2021learning} to encode the low-resolution image with $224 \times 224$, which has been pre-trained on 400 million image-text pairs sourced from the open-world data, thereby embedding extensive world knowledge into its pretrained weights. To preserve its generalization ability, we keep it frozen during the whole training procedure. The output sequence~$\mathit{V}_l$ is composed of 256 tokens, each with 1024 dimension.

\textbf{Projections.} The projections are designed to align visual tokens with the input token dimension of the subsequent large language model~\cite{chiang2023vicuna}. 
For the high-resolution feature map $\mathit{V}_h$, due to the limitation of input text length, we employ a 2D convolutional layer with a kernel size of 3 and a stride of 2, and then flatten it into $\frac{\mathrm{H}}{64} \times \frac{\mathrm{W}}{64}$ tokens, denoted as $\mathit{\hat{V}}_h$.
For the low-resolution visual features $\mathit{V}_l$, inspired from the paradigm of advanced LVLMs~\cite{liu2023visual,zhu2023minigpt}, we adopt a linear layer to project visual tokens, denoted as $\mathit{\hat{V}}_l$. 

\textbf{Concept Synergy.} Given the massive visual tokens and the embedding of textual instruction $\mathrm{Q}$, we utilize Vicuna-7B~\cite{chiang2023vicuna} as LLM to generate its response. 
Taking into account the discrepancy of table perception and comprehension tasks, we introduce \textit{meditative tokens} $\mathrm{M}$ to implement the concept synergy for the LLM, which adaptively enable different region of visual tokens and understand the intentions of specific task question. Finally, we construct the whole input sequence as $\mathit{X}$ = [$\mathrm{Q}$, \texttt{<IMG\_S>} ;$\mathit{\hat{V}}_l$ ; \texttt{<IMG\_SEP>} ; $\mathit{\hat{V}}_h$ ; \texttt{<IMG\_E>} ; $\mathrm{M}]$, where $[;]$ means the concatenation operation.
\texttt{<IMG\_S>}, \texttt{<IMG\_E>} and \texttt{<IMG\_SEP>} are learnable special tokens, that denote the start and end of visual tokens as well as the separation of different resolution tokens, respectively. 

\textbf{Objective.} Since TabPedia is trained to predict the next tokens like other LLMs, it is optimized by maximizing the likelihood of prediction loss at training time.

\subsection{Pre-training}
To enable the capable of vision encoders to capture text-rich information from high-resolution images and aligning embedding space with the large language model~\cite{chiang2023vicuna}, we first perform extensive text-aware pre-training. 
As shown in Fig.~\ref{framework}, we jointly optimize the high-resolution visual encoder with both projectors, while freezing the large language model and low-resolution vision encoder.
Specifically, followed by~\cite{feng2023docpedia}, our pre-training procedure involves a variety of perception tasks, \textit{i.e.}, text detection~\cite{liao2020real}, recognition~\cite{wang2011end}, spotting~\cite{liu2018fots}, long-text reading~\cite{kim2022ocr} and image captioning~\cite{hossain2019comprehensive}.
The first four tasks focuses on the various document images, while the last one targets natural scene images. These comprehensive tasks endow the vision encoders of TabPedia to effectively perceive textual and visual information from both document and natural scene images. 
More detailed pre-training settings about dataset and experiment could be referred to~\cite{feng2023docpedia}.

\subsection{Table-aware Fine-tuning}
Through pre-training, TabPedia could well understand text and structure of diverse document images but cannot follow instructions to perform different table understanding tasks.
In order to enhance the model capability of instruction following, we \textit{first} construct a large-scale dataset for visual table understanding. 
We will elaborate on the dataset construction in the Sec.~\ref{Sec.4}.
Based on this dataset, we introduce four table-related tasks, \textit{i.e.}, TD~\cite{smock2022pubtables}, TSR~\cite{smock2022pubtables,zheng2021global,zhong2020image}, TQ and TQA~\cite{zheng2021global,smock2022pubtables,pasupat2015compositional,chen2019tabfact} to simultaneously cultivate the perception and comprehension capabilities.
In this stage, we further unfreeze the LLM and fine-tune the entire framework except the low-resolution vision encoder.

\begin{table}[t]
	\begin{minipage}[l]{0.47\linewidth}
		\footnotesize
		\tabcolsep=10pt
		\tabcaption{Summary of training data statistics in the fine-tuning stage. }
		\label{Tab:summary}
		\renewcommand\arraystretch{1.04}
		\resizebox{\linewidth}{!}{
			\begin{tabular}{cccc}	
				\toprule
				\textbf{Dataset} & \textbf{Subset} & \textbf{Task} & \textbf{Num} \\  \midrule
                    \multirow{3}{*}{PubTab1M} & PubTab1M-Det & TD & 460k \\
                    & PubTab1M-Str & TSR,TQA & 759k \\
                    & PubTab1M-Syn & TQ & 381k \\
                    \midrule
                    FinTabNet & -- & TSR,TQA & 78k \\
                    \midrule
                    PubTabNet & -- & TSR & 434k \\
                    \midrule
                    WTQ & -- & TQA & 1k \\
                    \midrule
                    TabFact & -- & TQA & 9k \\
				\bottomrule
			\end{tabular}
		}
	\end{minipage}
        \hspace{0.4em}
        \begin{minipage}[r]{0.47\linewidth}
		\footnotesize
		\tabcolsep=10pt
		\tabcaption{Different task types and their instruction examples.}
		\label{Tab:Example}
		\renewcommand\arraystretch{1.04}
		\resizebox{\linewidth}{!}{
			\begin{tabular}{c|p{6cm}<{\centering}}	
				\toprule
				\textbf{Task} & \textbf{Example} \\        \midrule
                    TD & ``Give me the areas where table element's locations in this picture.'' \\
                    \midrule
                    TSR & ``Parse the structural information of the cropped table in this picture.'' \\
                    \midrule
                    TQ & ``Parse the table structure within the region [0.095, 0.673, 0.869, 0.851] in this picture.'' \\
                    \midrule
                    TQA & ``What was the lowest stock price in the fourth quarter of 2010?'' \\ 
				\bottomrule
			\end{tabular}
		}
	\end{minipage}
        \vspace{-1.5em}
\end{table}

\section{Dataset Construction}
\label{Sec.4}
In this section, we aim to introduce the collected instruction following dataset.
The entire data is derived from five public datasets, including PubTab1M~\cite{smock2022pubtables}, FinTabNet~\cite{zheng2021global}, PubTabNet~\cite{zhong2020image}, WikiTableQuestions~(WTQ)~\cite{pasupat2015compositional} and TabFact~\cite{chen2019tabfact}.
Among them, PubTab1M~\cite{smock2022pubtables} contains two subsets, \textit{i.e.}, PubTab1M-Detection~(PubTab1M-Det) and PubTab1M-Structure~(PubTab1M-Str).
Moreover, since the table images in PubTab1M-Str are cropped from PubTab1M-Det, we transform the annotations of the table structure in PubTab1M-Str into the original images and synthesize a new subset PubTab1M-Syn, which could be utilized for TQ task.
The statistical data are summarized in Tab.~\ref{Tab:summary}. 
To ensure the instruction diversity, we generate multiple instructions for each task using GPT3.5~\cite{brown2020language}. In Tab.~\ref{Tab:Example}, we display one exemplar about user's question for each table task.
We will provide a detailed exposition of them in the following.

\textbf{Table Detection~(TD).} As a fundamental task, TD task targets to detect all table locations in a document image. Previous methods~\cite{Prasad_CascadeTabNet_2020,smock2022pubtables,siddiqui2018decnt} mainly utilize DETR~\cite{carion2020end} or variants of R-CNN~\cite{girshick2015fast,ren2015faster,he2017mask} to predict numerous overlapping bboxes, that inevitably needs complex post-processing, such as non-maximization suppression~(NMS), to generate final results. In contrast, we employ LLM to directly generate the locations of instance tables in the format of ``[x1, y1, x2, y2]'', where x1, y1, x2, y2 represent the normalized coordinates of the top-left and bottom-right of the corresponding bbox. 
Moreover, to facilitate detection results for multiple tables, we split multiple table positions with the special symbol ``\textbackslash n'' in the output response. 
We adopt PubTab1M-Det~\cite{smock2022pubtables} to perform TD task, where images are collected from PDF documents with different scale and rotation types of tables.

\textbf{Table Structure Recognition~(TSR).} The TSR targets to parse table structure in terms of rows, columns and cells. HTML and Markdown codes are mainly two kinds of text sequences used to represent a table. HTML could represent all kinds of tables, with or without cells spanning multiple rows and grids, but they contain massive markup grammars~\textit{i.e.}, ``\texttt{<div></div>}'' and ``\texttt{<td></td>}'', resulting in excessively lengthy output responses. 
Compared with HTML, Markdown represents a table more succinctly, but it cannot represent cells spanning multiple rows or columns. 
By weighing the simplicity of the output and the completeness of the table parsing, we propose a canonical table structure representation based on the detection format. Inspired by~\cite{smock2022pubtables}, we jointly adopt five object classes to model TSR, including \textit{table column}, \textit{table row}, \textit{table column header}, \textit{table projected row header} and \textit{table spanning cell}.
To better understanding, we display a representative sample in Appendix~\ref{appendix:annotation}.
Taking into account the serialized output of the LLM, we represent the table structure with a series of ``[object] [x1, y1, x2, y2]'', which are also separated by ``\textbackslash n''.
Notably, we standardize the order of the output objects to ensure uniqueness of the table parsing results. 

We select the PubTab1M-Str~\cite{smock2022pubtables}, FinTabNet~\cite{zheng2021global} and PubTabNet~\cite{zhong2020image} to support the TSR task, where tables are collected from scientific and financial articles. 
These datasets contain pairs of table images and HTML annotations. 
We convert HTML codes into our designed annotation format using the pre-processing tool offered by~\cite{smock2022pubtables}.

\textbf{Table Querying~(TQ).} Different from recognizing table structure from the cropped table-centric images in TSR task, the TQ task directly parses the table from the original document image based on the given table location. 
This task is more challenging due to the degradation of the table's resolution and the interference of other document contents around it.
Moreover, this task could potentially be combined with TD task to enable automatic parsing of all table structure information in original images. 
Therefore, we introduce this task to fully unlock the comprehension capabilities of large language models for visual table understanding. 
For the annotation of table parsing, we adopt the same 
format as TSR.
Since there is no readily available dataset, we synthesize a large amount of available data based on the annotations from PubTab1M~\cite{smock2022pubtables}, namely PubTab1M-Syn. 

\noindent \textbf{Table Question Answering~(TQA).} 
TQA aims to provide precise answers through table understanding and reasoning. For both public TQA datasets, \textit{i.e.}, WTQ~\cite{pasupat2015compositional} and TabFact~\cite{chen2019tabfact}, the table images are collected from wikipedia tables with pairs of content-related question and answer. Thus, we could directly apply these available data to support this task. 
However, the images of current TQA data are rendered from text-based tables with variations in background color and font size, resulting in poor generalization in real-world tables. In addition, the TQA data volume lags far behind other tasks. 
To alleviate these obstacles, we generate numerous TQA data with partial images in FinTabNet~\cite{zheng2021global} and PubTab1M~\cite{smock2022pubtables} by employing the powerful multi-modal understanding capabilities of Gemini Pro~\cite{reid2024gemini}. 
We provide more detailed descriptions of the procedure in the Appendix~\ref{appendix:Collection}

To better evaluate TQA performance of various models on real-world table images, we build a complex TQA dataset~(ComTQA) based on test set of FinTabNet~\cite{zheng2021global} and PubTab1M~\cite{smock2022pubtables}.
Compared to WTQ and TabFact, ComTQA has more challenging questions, such as multiple answers, mathematical calculations, and logical reasoning.
In total, we annotate $\sim$9k high-quality QA pairs from $\sim$1.5k images by expert annotation.
More statistics about ComTQA could be found in the Appendix~\ref{appendix:ComTQA}.

\section{Experiment}
\label{Experiment}
\subsection{Implementation Details}
\textbf{Parameter Settings.} For the hyper-parameters in model design, the number of meditative tokens is set to 256. The max length of text sequence is set to 4000 to satisfy task requirements. 
To implement TabPedia, we adopt a cosine schedule with one-cycle learning rate strategy~\cite{smith2019super}.
In the pre-training phase, the learning rate warms up in the first 2\% of the training process and then decreases from the peak rate~(1e-3) with batch sizes of 64.
In the fine-tuning phase, we set the peak learning rate as 5e-6 with batch sizes of 16.
We employ the AdamW optimizer~\cite{loshchilov2018decoupled} in both phases.
All experiments are implemented by PyTorch~\cite{paszke2019pytorch} and trained on 16$\times$ A100 GPUs. 

\textbf{Datasets.}
In order to comprehensively evaluate the capability of TabPedia, we employ multiple benchmarks for each task.
For performance assessment, we set the temperature parameter as 0.2 in both quantitative and qualitative evaluations.
For TD task, PubTab1M-Det~\cite{smock2022pubtables} contains 57,125 images for testing.
For TSR task, FinTabNet~\cite{zheng2021global}, PubTabNet~\cite{zhong2020image} and PubTab1M-Str~\cite{smock2022pubtables} are adopted for evaluation with 9,289, 9,115 and 93,834 testing samples, respectively.
For TQ task, the synthetic dataset PubTab1M-Syn~\cite{smock2022pubtables} also provides 47,186 samples for testing.
For TQA task, WTQ~\cite{pasupat2015compositional}, TabFact~\cite{chen2019tabfact} and our annotated ComTQA contain 4,343, 12,722 and 9,070 QA pairs, respectively. 

\textbf{Evaluation Metrics.} 
For TD task, we report the results with object detection metrics, including precision, recall and f1-score with IoU@0.75.
For both TSR and TQ tasks, we utilize Structure Tree-EditDistance-based Similarity (S-TEDS)~\cite{zhong2020image}, which evaluates table similarity of structural aspects in HTML format. The metric represents the HTML table as a tree, and the TEDS score is computed through the tree-edit distance between the ground truth and predicted trees. In order to convert the results of TabPedia into HTML format, we employ the post-processing algorithm provided by~\cite{smock2022pubtables}.
Moreover, we report the recently proposed GriTS metrics~\cite{smock2023grits} for PubTab1M-Str to align its original metric. 
Different from S-TEDS, GriTS represents tables as matrices, better capturing the two-dimensional structure and the orders of cells in a table. Further, GriTS enables TSR to be assessed from multiple perspectives, with $\mathrm{GriTS_{Top}}$ measuring cell topology recognition, $\mathrm{GriTS_{Cont}}$ measuring cell content recognition, and $\mathrm{GriTS_{Loc}}$ measuring cell location recognition.
For TQA task, we adopt the accuracy metric where the response generated by the model is judged correct if it contains the string present in the ground truth~\cite{liu2023hidden}.

\begin{table}[t]
    \begin{minipage}[c]{0.56\linewidth}
    \footnotesize
    \tabcolsep=8pt
    \tabcaption{Comparison with the existing best table detection model TATR~\cite{smock2022pubtables}. NMS denotes Non-Maximum Suppression.}
    \label{Tab:TD}
    \vspace{0.5em}
    \renewcommand\arraystretch{1.04}
    \resizebox{\linewidth}{!}{
        \begin{tabular}{cccccc}	
            \toprule
            \multirow{2}{*}{\textbf{Method}} & \multirow{2}{*}{\textbf{Backbone}} &  \multirow{2}{*}{\textbf{NMS}} & \multicolumn{3}{c}{IoU@0.75}\\
            \cmidrule{4-6} 
       & & & \textbf{Precision} & \textbf{Recall} & \textbf{F1} \\
        \midrule
        \multirow{2}{*}{TATR~\cite{smock2022pubtables}} & Faster R-CNN & \Checkmark &92.7 & 86.6 & 89.5 \\
        & DETR & \Checkmark & \textbf{98.8} & 98.1 & \textbf{98.4} \\
        \midrule
        \textbf{TabPedia} & LVLM & \XSolidBrush & 98.5 & \textbf{98.4} & \textbf{98.4} \\
            \bottomrule
        \end{tabular}
    }	
    \end{minipage} 
    \hspace{0.5em}
    \begin{minipage}[c]{0.37\linewidth}
    \footnotesize
    \tabcolsep=3pt
    \tabcaption{Comparison with end-to-end TSR methods on two datasets. ``$\ast$'' represents the results reported by~\cite{wan2024omniparser}.}
    \label{Tab:TSR}
    \renewcommand\arraystretch{1.04}
    \resizebox{\linewidth}{!}{
        \begin{tabular}{cccc}	
            \toprule
            \multirow{2}{*}{\textbf{Method}} & \multirow{2}{*}{\textbf{Input Size}} & \multicolumn{1}{c}{\textbf{PubTabNet}} & \multicolumn{1}{c}{\textbf{FinTabNet}} \\
            \cmidrule(lr){3-3} \cmidrule(lr){4-4}
       & & \textbf{S-TEDS} & \textbf{S-TEDS} \\
        \midrule
        Donut~\cite{kim2022ocr}$^\ast$ & 1,280 & 25.28 & 30.66 \\
        EDD~\cite{zhong2020image} & 512 & 89.90 & 90.60 \\
        OmniParser~\cite{wan2024omniparser} & 1,024 & 90.45 & 91.55 \\
        \midrule
        \textbf{TabPedia} & 2,560 & \textbf{95.41} & \textbf{95.11} \\
        \bottomrule
        \end{tabular}
    }	
    \end{minipage} 
    \vspace{-0.5em}
\end{table}

\begin{table}[t]
    \begin{minipage}[c]{0.56\linewidth}
        \caption{Quantitative results on two subsets of PubTab1M~\cite{smock2022pubtables}, including PubTab1M-Str and PubTab1M-Syn.}
        \vspace{-0.3em}
        \begin{subtable}{\linewidth}
            \footnotesize
            \tabcolsep=3pt
            \tabcaption{Comparison with the task-specific model, TATR~\cite{smock2022pubtables} on TSR task. ``Cropped'' denotes utilizing cropped table-centric images.}
            \label{Tab:TD_TQ}
            \renewcommand\arraystretch{1.04}
            \resizebox{\linewidth}{!}{
                \begin{tabular}{cccccccc}	
                    \toprule
                    \multirow{2}{*}{\textbf{Method}} & \multirow{2}{*}{\textbf{Backbone}} & \multirow{2}{*}{\textbf{Image}} & \multirow{2}{*}{\textbf{NMS}} & \multicolumn{4}{c}{\textbf{PubTab1M-Str}}\\
                    \cmidrule{5-8} 
               & & & & $\mathbf{GriTS_{Top}}$ & $\mathbf{GriTS_{Cont}}$ & $\mathbf{GriTS_{Loc}}$ & \textbf{S-TEDS}\\
                \midrule
                \multirow{2}{*}{TATR~\cite{smock2022pubtables}} & Faster R-CNN &  Cropped & \Checkmark &86.16 & 85.38 & 72.11 & -- \\
                 & DETR  &  Cropped  & \Checkmark & \textbf{98.46} & \textbf{97.81} & \textbf{97.81} & \textbf{97.65} \\
                \midrule
                \textbf{TabPedia}~(TSR) & LVLM & Cropped & \XSolidBrush & 96.52 & 96.73 & 95.54 & 95.66 \\
                \bottomrule
                \end{tabular}
            }
        \end{subtable}
        \begin{subtable}{\linewidth}
            \footnotesize
            \tabcolsep=3pt
            \tabcaption{Quantitative results on both TQ and TD+TQ tasks.}
            \vspace{-0.3em}
            \label{tab:TQ}
            \renewcommand\arraystretch{1.04}
            \resizebox{\linewidth}{!}{
                \begin{tabular}{cccccccc}	
                    \toprule
                    \multirow{2}{*}{\textbf{Method}} & \multirow{2}{*}{\textbf{Image}} & \multirow{2}{*}{\textbf{NMS}} & \multirow{2}{*}{\textbf{Task}} & \multicolumn{4}{c}{\textbf{PubTab1M-Syn}}\\
                    \cmidrule{5-8} 
               & & & & $\mathbf{GriTS_{Top}}$ & $\mathbf{GriTS_{Cont}}$ & $\mathbf{GriTS_{Loc}}$ & \textbf{S-TEDS}\\
                \midrule
                \multirow{2}{*}{\textbf{TabPedia}}& \multirow{2}{*}{\textbf{Raw}} & \multirow{2}{*}{\textbf{\XSolidBrush}} & TQ & 96.04 & 96.23 & 94.95 & 95.07\\
                 & & & TD+TQ & 94.54 & 94.63 & 93.25 & 93.38 \\
                \bottomrule
                \end{tabular}
            }
        \end{subtable}
   \end{minipage} 
   \hspace{0.5em}
   \begin{minipage}[c]{0.38\linewidth}
        \footnotesize
        \tabcolsep=3pt
        \tabcaption{Comparison with existing LVLMs on TQA task. ``$\ast$'' denotes the results obtained through the open-source checkpoint or API of the closed-source model. ComTQA is our released new benchmark. The second best methods are underlined.}
        \vspace{-0.3em}
        \label{Tab:TQA}
        \renewcommand\arraystretch{1.04}
        \resizebox{\linewidth}{!}{
            \begin{tabular}{ccccc}	
                \toprule
                \multirow{2}{*}{\textbf{Method}} & \multirow{2}{*}{\textbf{Input Size}} & \multicolumn{1}{c}{\textbf{WTQ}} & \multicolumn{1}{c}{\textbf{TabFact}} & \multicolumn{1}{c}{\textbf{ComTQA}}\\
                \cmidrule(lr){3-3} \cmidrule(lr){4-4} \cmidrule(lr){5-5}
           & & \textbf{Acc} & \textbf{Acc} & \textbf{Acc}\\
            \midrule
            TextMonkey~\cite{liu2024textmonkey} & 896 & 37.9 & 53.6 & 13.9$^\ast$\\
            Monkey~\cite{li2023monkey} & 896 & 25.3$^\ast$ & 49.8 & --\\
            Cogagent~\cite{hong2023cogagent} & 1,120 & 30.2$^\ast$ & 51.7$^\ast$ & --\\
            DocOwl 1.5~\cite{hu2024mplug} & 1,344 & 39.8 & \textbf{80.4} & 18.5$^\ast$ \\
            GPT4V~\cite{gpt4v} & 645 & \underline{45.5}$^\ast$ & 69.3$^\ast$ & 27.2$^\ast$ \\
            Gemini Pro~\cite{reid2024gemini} & 659 & 32.3$^\ast$ & 67.9$^\ast$ & \underline{29.3}$^\ast$ \\
            Xcomposer2~\cite{dong2024internlm} & 511 & 28.7 & 62.3 & -- \\
            \midrule
            \textbf{TabPedia} & 2,560 & \textbf{47.8} & \underline{71.3} & \textbf{53.5} \\
            \bottomrule
            \end{tabular}
        }	
    \end{minipage}   
    \vspace{-1.0em}
\end{table}

\subsection{Quantitative Results}
We conduct quantitative evaluations of current state-of-the-art methods for specific tasks in perception and comprehension, comparing them to our proposed TabPedia.

\textbf{Evaluation on TD.} 
In Tab.~\ref{Tab:TD}, we compare TabPedia with the previous state-of-the-art method, TATR~\cite{smock2022pubtables}.
TATR performs the table detection with two classic visual detection backbones, \textit{i.e,} DETR~\cite{carion2020end} and Faster R-CNN~\cite{ren2015faster}.
Compared with them, TabPedia outperforms Faster R-CNN with a notable margin and achieves competitive performance with DETR.
Notably, since TabPedia directly generates the independent locations of instance tables without densely overlapped bboxes, there are no extra post-processing operations involved, \textit{i.e.}, Non-Maximum Suppression~(NMS).
This advantage could enable TabPedia to perform more complex table understanding, such as parsing all tables by combining TD and TQ tasks. 

\textbf{Evaluation on TSR.} Tab.~\ref{Tab:TSR} reports the performance of TSR task compared to end-to-end TSR models on PubTabNet and FinTabNet datasets. Specifically, the OCR-free model Donut~\cite{kim2022ocr} is fine-tuned for TSR with the official default training configuration. 
Although OmniParser~\cite{wan2024omniparser} integrates multiple visually-situated text parsing tasks into a unified framework, it adopts three isolated decoders to perform different tasks.
Compared with OmniParser, TabPedia consistently surpasses it with 4.96\% and 3.56\% S-TEDS on both datasets, respectively.
In Tab.~\ref{Tab:TD_TQ}, TATR as the task-specific method, shows high performance with the DETR architecture.
Our proposed TabPedia, a generic model for tasks involving both perception and comprehension, still achieves comparable performance without the need for complex post-processing.
These results highlight the exceptional capability of TabPedia.

\textbf{Evaluation on TQ.} As a new and unexplored task, the TQ task aims to parse table structures with the specific location directly from the raw image without additional cropping.
In the first row of Tab.~\ref{tab:TQ}, we provide a strong baseline with 96.04\% and 95.07\% on $\mathrm{GriTS}_{\mathrm{Top}}$ and S-TEDS, respectively, which nearly reaches the same performance as parsing from the cropped images under the interference of the document content around the table. 
Furthermore, we integrate both TD and TQ tasks in the form of multi-round dialogue, which endows TabPedia to directly parse all existing tables in a document image. We report the final result in the second row of Tab.~\ref{tab:TQ}.
These impressive results demonstrate that TabPedia has the potential to enable more holistic table understanding. 

\textbf{Evaluation on TQA.} Due to the complex structure of tables and the dense text, the understanding of the table contents remains a challenging issue. To thoroughly evaluate the performance of the understanding of table content and structure, we adopt two public benchmarks, \textit{i.e.}, WTQ~\cite{pasupat2015compositional} and TabFact~\cite{chen2019tabfact}, and our collected dataset ComTQA, as shown in Tab.~\ref{Tab:TQA}. 
On the WTQ and TabFact, TabPedia achieves promising performance among the open and close sources LVLMs.
In contrast to existing benchmarks, ComTQA contains real-world table images with more complex questions. 
It is observed that current LVLMs show poor performance due to the incomplete understanding of real-world table structures.
Compared with them, TabPedia achieves the optimal result with a notable margin, which demonstrates the effectiveness of jointly learning perception and comprehension tasks. 

\begin{figure}[t]
	\centering
	\includegraphics[width=0.97\textwidth]{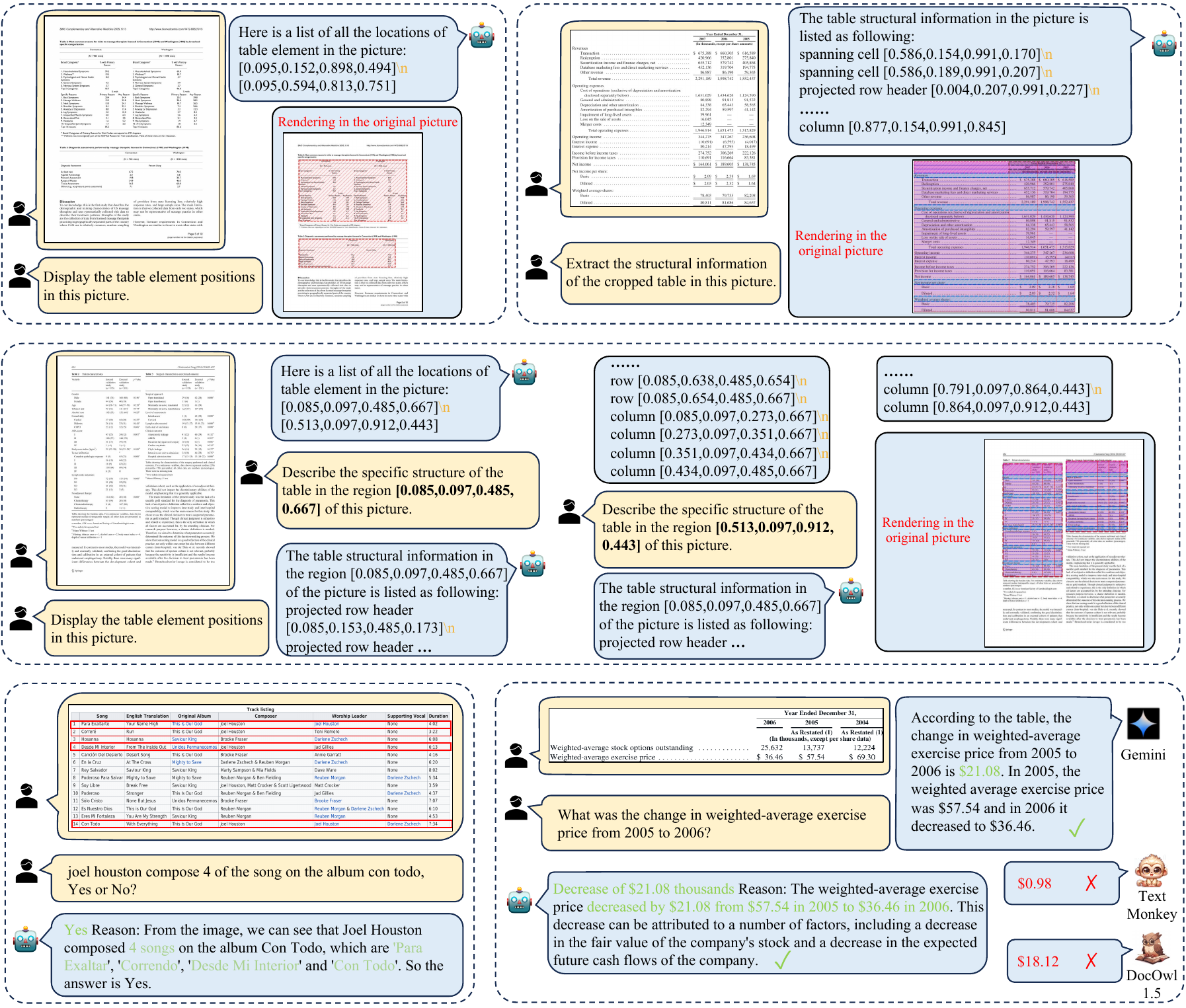} 
	\vspace{-0.5em}
	\caption{Qualitative results of TabPedia on diverse tasks. The first row shows its perception capability on both TD and TSR tasks. The second row further exhibits TabPedia's powerful ability by employing multiple instructions of different tasks. The bottom row showcases TabPedia's accurate responses based on intricate contents in visual tables. Zoom in for best view.}
	\label{Qualitative}
 \vspace{-1.5em}
\end{figure}

\subsection{Qualitative Results}
\vspace{-0.5em}
We further conduct qualitative evaluation on TabPedia's perception and comprehension capabilities.
Firstly, we show the perception capability of TabPedia with solely TD and TSR tasks, as illustrated in the first row of Fig.~\ref{Qualitative}.
TabPedia accurately generates reliable and formatted results, which are rendered to the original image for better observation. 
Secondly, TabPedia performs a complex task to directly parsing all table structure information in a document image by integrating instructions of TD and TQ tasks within a multi-round dialogue.
As shown in the second row of Fig.~\ref{Qualitative}, the example indicates that TabPedia is capable of exploring more holistic visual table understanding. 
In the last row, we display the table comprehensive capability of TabPedia.
It is observed that the response not only contains concise and reliable answer, but also provides the specific contents in the table to support its answer.
Especially, TabPedia even acquires certain math calculation ability to capture the connections among table contents, as shown in the bottom right example in Fig.~\ref{Qualitative}. 
These results demonstrate Tabpedia’s powerful multimodal comprehension
capabilities. 
We also display more visualization results in the Appendix~\ref{More_Qualitative}.

\subsection{Ablation Studies}
In this section, we conduct ablation studies to validate the effectiveness of core settings and components in TabPedia. 
All experiments are conducted on three datasets across three tasks: PubTab1M-Det~\cite{smock2022pubtables}, FinTabNet~\cite{zheng2021global} and WTQ~\cite{pasupat2015compositional}.

\textbf{Necessity of Meditative Tokens.} In Tab.~\ref{Abla:token}, we conduct the experiment to investigate the impact of adding meditative tokens in TabPedia. 
It is observed that adding meditative tokens significantly improves TabPedia's capabilities of table perception and comprehension. 

\textbf{What Information Matters for Meditative Tokens?} We sample 100 test cases for each task and report the averaged numeric importance of high- and low-resolution vision tokens when they are attended by the meditative tokens for different tasks in the Tab.~\ref{Abla:visual token weights}. Specifically, for the various VTU tasks, we calculate the averaged attention scores (across all layers and attention heads) from the LLM decoder, which indicates the extent to which the meditative tokens focus on either high- or low-resolution visual tokens.
For the TSR and TQ tasks, the meditative tokens pay significantly more attention to the high-resolution visual encoder tokens. We attribute this to the fact that both tasks require more fine-grained visual information to be "deliberated" in order to construct the dense table structure. In contrast, for the TD and TQA tasks, the two visual encoders contribute almost equally to the information attended to by the meditative tokens, validating the importance of both vision encoders for these tasks.

\begin{wraptable}{r}{0.44\textwidth}
    \footnotesize
    \tabcolsep=6pt
    \tabcaption{Contributions of different tokens.}
    \label{Abla:information weights}
    \renewcommand\arraystretch{1.02}
    \resizebox{\linewidth}{!}{
        \begin{tabular}{@{}cccc@{}}	
            \toprule
            \multirow{1}{*}{\tabincell{c}{Task}} & \multicolumn{1}{c}{\tabincell{c}{Meditative \\ tokens}} & \multicolumn{1}{c}{\tabincell{c}{High-res \\ visual tokens}} & \multicolumn{1}{c}{\tabincell{c}{Low-res \\ visual tokens}} \\
        \midrule
        TD & 0.65 & 0.16 & 0.19 \\
        TSR & 0.64 & 0.12 & 0.24 \\
        TQ & 0.71  & 0.11 & 0.19 \\
        TQA & 0.56 & 0.18 & 0.25 \\
        \bottomrule
        \end{tabular}
    }
\end{wraptable}
\textbf{Contributions of Different Tokens.} In the Tab.~\ref{Abla:information weights}, we calculate the averaged scores of the TabPedia-generated answers with respect to meditative tokens, high-resolution visual tokens, and low-resolution visual tokens across all the attention maps from the LLM, respectively. One can observe that the meditative tokens contribute the most information to the generation of satisfactory answers, which demonstrates that the proposed meditative tokens are indispensable and effective. 
We also provide a detailed analysis of the attention map of meditative tokens in Fig.~\ref{fig:atten_map} of Appendix.~\ref{More_Qualitative}.

\begin{table}[t]
    \begin{minipage}[c]{0.5\linewidth}
    \footnotesize
    \tabcolsep=6pt
    \tabcaption{Impact of meditative tokens in TabPedia.}
    \label{Abla:token}
    \renewcommand\arraystretch{1.04}
    \resizebox{\linewidth}{!}{
        \begin{tabular}{@{}cccc@{}}	
            \toprule
            \multirow{2}{*}[-0.3em]{\tabincell{c}{meditative\\token}} & \multicolumn{1}{c}{\textbf{PubTab1M-Det}} & \multicolumn{1}{c}{\textbf{FinTabNet}} & \multicolumn{1}{c}{\textbf{WTQ}} \\
            \cmidrule(lr){2-2} \cmidrule(lr){3-3} \cmidrule(lr){4-4} 
       & \textbf{Precision} & \textbf{S-TEDS} & \textbf{Acc} \\
        \midrule
         \usym{2715} & 93.5 & 92.17 & 43.2 \\
        \checkmark & \textbf{98.5} & \textbf{95.11} & \textbf{47.8}\\
        \bottomrule
        \end{tabular}
    }	
    \end{minipage} 
    \hspace{0.5em}
    \begin{minipage}[c]{0.45\linewidth}
    \footnotesize
    \tabcolsep=3pt
    \tabcaption{Impact of different training strategies on the low-resolution vision encoder.}
    \label{Abla:visual token weights}
    \renewcommand\arraystretch{1.02}
    \resizebox{\linewidth}{!}{
        \begin{tabular}{ccc}	
            \toprule
            \multirow{1}{*}{\tabincell{c}{Task}} & \multicolumn{1}{c}{\textbf{High-res visual tokens}} & \multicolumn{1}{c}{\textbf{Low-res visual tokens}} \\
        \midrule
        TD & 0.49 & 0.51 \\
        TSR & 0.71 & 0.29 \\
        TQ & 0.73  & 0.27  \\
        TQA & 0.51 & 0.49 \\
        \bottomrule
        \end{tabular}
    }
    \end{minipage} 
    \vspace{-1.0em}
\end{table}

\textbf{Impact of Dual Vision Encoders.} 
As shown in Table~\ref{Abla:encoders}, we explore the impact of different vision encoders that capture global and local information from input images at various resolutions. The high-resolution encoder extracts intricate details from text-rich images, outperforming the low-resolution encoder, which struggles with nuanced visual representations in complex document images.
Different tasks may require distinct visual cues, so dual vision encoders offer flexibility. For instance, TQA tasks need detailed table information, while TSR tasks depend on global layout. The low-resolution encoder provides comprehensive layout insights, complementing the high-resolution encoder's limited receptive field.
Our results demonstrate that combining both encoders enhances the extraction of structural and content-related details from tables, improving perception and comprehension tasks.

\textbf{Frozen vs. Unfrozen Low-Resolution Vision Encoder.} We further investigate different training strategies in terms of the low-resolution vision encoder. As shown in Tab.~\ref{Abla:low-res enc}, it is observed that no significant performance improvement but with longer training time consumption by unfreezing it, which is in line with the conclusion in the pioneering work~\cite{huangfroster}. Besides, we suppose the encoder frozen can serve as a regularization, facilitating the extraction of layout information and alleviating potential overfitting problems, as well as more stable training. To strike the trade-off between computational consumption and performance, we thus freeze the low-resolution vision encoder during training. 

\begin{table}[t]
    \begin{minipage}[c]{0.45\linewidth}
    \footnotesize
    \tabcolsep=6pt
    \tabcaption{Impact of different training strategies on low-resolution vision encoder.}
    \label{Abla:low-res enc}
    \renewcommand\arraystretch{1.04}
    \resizebox{\linewidth}{!}{
        \begin{tabular}{@{}cccc@{}}	
            \toprule
            \multirow{2}{*}[-0.3em]{\tabincell{c}{Low-Res \\ Encoder}} & \multicolumn{1}{c}{\textbf{PubTab1M-Det}} & \multicolumn{1}{c}{\textbf{FinTabNet}} & \multicolumn{1}{c}{\textbf{WTQ}} \\
            \cmidrule(lr){2-2} \cmidrule(lr){3-3} \cmidrule(lr){4-4} 
       & \textbf{Precision} & \textbf{S-TEDS} & \textbf{Acc} \\
        \midrule
         frozen & \textbf{98.5} & 95.11 & \textbf{47.8} \\
        unfrozen & 98.4 & \textbf{95.11} & 46.4 \\
        \bottomrule
        \end{tabular}
    }	
    \end{minipage} 
    \hspace{0.5em}
    \begin{minipage}[c]{0.5\linewidth}
    \footnotesize
    \tabcolsep=4pt
    \tabcaption{Impact of dual vision encoders.}
    \label{Abla:encoders}
    \renewcommand\arraystretch{1.04}
    \resizebox{\linewidth}{!}{
        \begin{tabular}{@{}ccccc@{}}	
            \toprule
            \multirow{2}{*}[-0.3em]{\tabincell{c}{High-Res\\Encoder}} & \multirow{2}{*}[-0.3em]{\tabincell{c}{Low-Res\\Encoder}} & \multicolumn{1}{c}{\textbf{PubTab1M-Det}} & \multicolumn{1}{c}{\textbf{FinTabNet}} & \multicolumn{1}{c}{\textbf{WTQ}} \\
            \cmidrule(lr){3-3} \cmidrule(lr){4-4} \cmidrule(lr){5-5}
       & & \textbf{Precision} & \textbf{S-TEDS} & \textbf{Acc} \\
        \midrule
        \checkmark & &  96.5 & 93.6 & 44.9 \\
        & \checkmark & 86.2 & 81.3 & 24.7 \\
        \checkmark & \checkmark & \textbf{98.5} & \textbf{95.11} & \textbf{47.8}\\
        \bottomrule
        \end{tabular}
    }	
    \end{minipage} 
\end{table}

\vspace{-0.5em}
\section{Limitation}
\vspace{-0.5em}
\label{limitation}
In this section, we discuss the limitations of our TabPedia.
Firstly, since we represent the table structure with regular rectangular boxes, TabPedia is currently not capable of accurately parsing structural information for twisted or distorted tables.
Secondly, all images in TQA datasets, including WTQ~\cite{pasupat2015compositional}, TabFact~\cite{chen2019tabfact} and ComTQA are dominated by tables. Therefore, TabPedia still lacks the capability to directly answer the table question with original document image. In addition, compared to parallel decoding algorithms such as DETR~\cite{carion2020end} and Faster R-CNN~\cite{ren2015faster}, it consumes longer decoding time. Meantime, certain algorithmic designs such as KV cache, flash attention, and hardware improvements can effectively improve inference efficiency. We believe that with the iterative development of large model technology, the inference efficiency of TabPedia can be significantly improved.

\vspace{-0.5em}
\section{Conclusion}
\vspace{-0.5em}
In this paper, we propose a novel large vision-language model to unify diverse visual table understanding tasks, namely TabPedia.
Specifically, we present a \textit{concept synergy} mechanism to seamlessly integrate diverse tasks and multi-source visual tokens embedded from dual vision encoders as \textit{concepts}.
This mechanism is implemented by introducing the \textit{meditative tokens} into the LLM.
Then, we fully leverage the capability of LLMs to effectively understand these concepts and generate accurate and plausible responses.
Extensive quantitative and qualitative experiments across various public benchmarks validate the effectiveness of our TabPedia.
To further investigate the potential of TabPedia, we establish a challenging table VQA dataset, ComTQA, featuring round 9,000 QA pairs.

\bibliography{main}
\bibliographystyle{unsrtnat}

\newpage

\appendix
\renewcommand\thesection{\Alph{section}} 
		\renewcommand\thesubsection{\Alph{section}.\arabic{subsection}} 
		\renewcommand\thefigure{\Alph{section}\arabic{figure}} 
		\renewcommand\thetable{\Alph{section}\arabic{table}} 
		\setcounter{section}{0}
		\setcounter{figure}{0}	
		\setcounter{table}{0}
\section{More details about TQA datasets}

\subsection{QA Pairs Generation}
\label{appendix:Collection}
\begin{figure}[!h]
	\centering
	\includegraphics[width=1.0\textwidth]{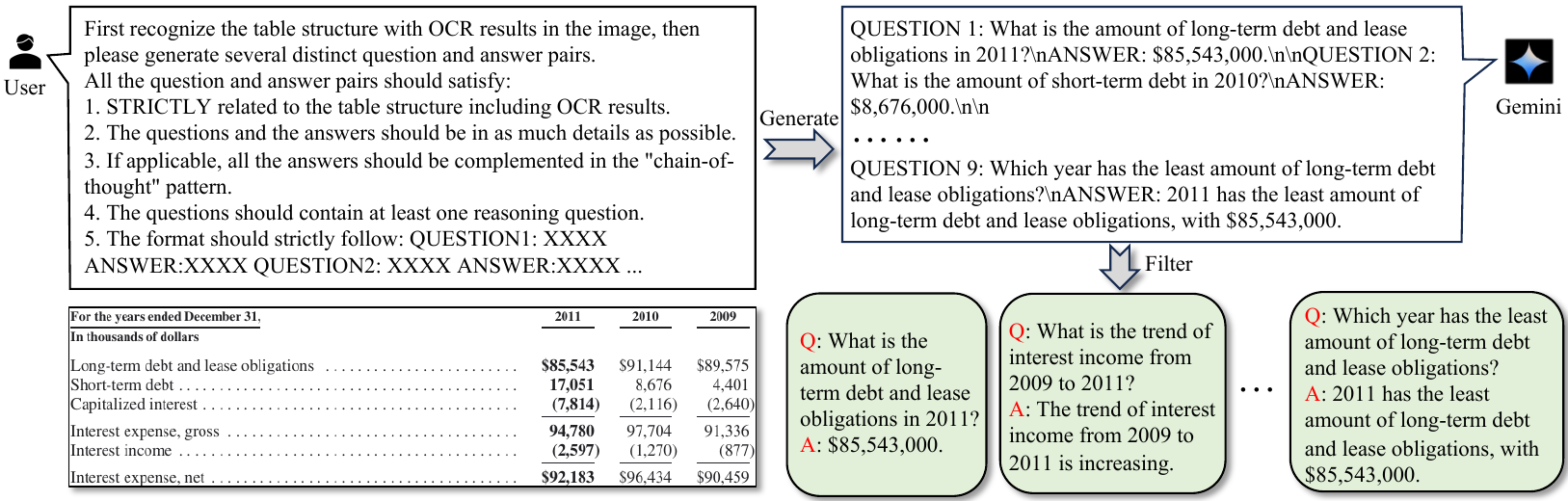} 
	\vspace{-1.0em}
	\caption{The illustration of an example for generating QA pairs with the powerful LVLM, Gemini Pro~\cite{reid2024gemini}. The prompt includes several key rules to ensure the response quality as much as possible.}
	\label{fig:collection}
\end{figure}

We depict the procedure of collecting QA pairs with an example in Fig.~\ref{fig:collection}. 
For input image, Gemini Pro~\cite{reid2024gemini} is prompted to first recognize the table structure with OCR results in the image, then generate several question and answer pairs according to OCR results.
In order to improve the reliability of the generated answers, we leverage various prompting techniques, \textit{i.e,} Chain-of-Thought and few-shot prompting.
According to the specific prompt, Gemini Pro will generate multiple QA pairs for each input image and return them in an agreed-upon format. 
After obtaining raw responses generated by Gemini Pro, we utilize the regularized matching algorithm and the special character filter in turn to extract available question and answer pairs.

\subsection{ComTQA Benchmark}
\label{appendix:ComTQA}
\begin{figure}[!h]
    \begin{minipage}[c]{0.5\linewidth}
    \footnotesize
    \tabcolsep=8pt
    \tabcaption{Statistics of ComTQA benchmark.}
    \label{appendix:statistic}
    \renewcommand\arraystretch{1.04}
    \resizebox{\linewidth}{!}{
        \begin{tabular}{@{}ccc|c@{}}	
            \toprule
        & \textbf{PubTab1M} & \textbf{FinTabNet} & \textbf{Total} \\
        \midrule
        \#images & 932 & 659 & 1,591 \\
        \#QA pairs & 6,232 & 2,838 & 9,070 \\
        \tabincell{c}{Avg. per image} & 6 & 4 & 5\\
        \bottomrule
        \end{tabular}
    }	
    \end{minipage} 
    \begin{minipage}[c]{0.5\linewidth}
        \centering
	\includegraphics[width=1.0\textwidth]{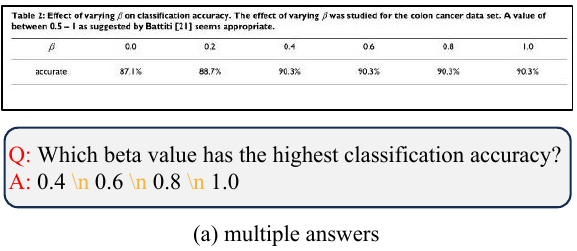} 
	\vspace{-1.0em}
    \end{minipage}
\end{figure}

\begin{figure}[!h]
     \centering
	\includegraphics[width=1.0\textwidth]{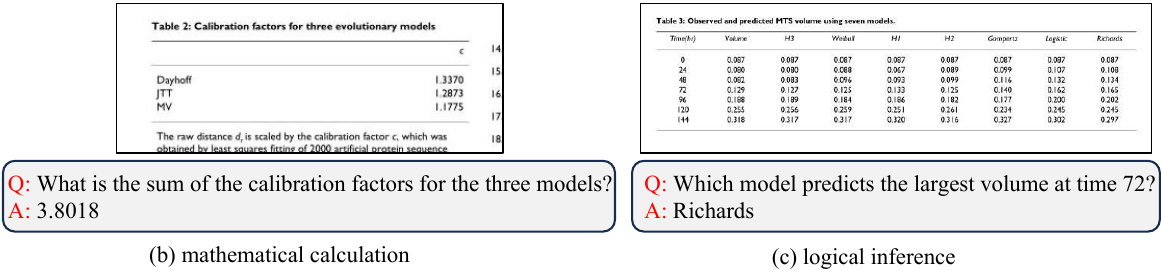} 
	\vspace{-1.0em}
	\caption{More visualization on ComTQA benchmark. We display several complex QA types, such as multiple answers, mathematical calculation and logical inference. Zoom in for best view.}
	\label{fig:com_tqa2}
\end{figure}

In Tab.~\ref{appendix:statistic}, we present the distribution of both data sources~\cite{smock2022pubtables,zheng2021global} within the ComTQA dataset.
Concretely, ComTQA comprises a total of 9,070 QA pairs across 1,591 images, averaging 5 questions per image.
Different from existing TQA benchmarks~\cite{chen2019tabfact,pasupat2015compositional}, ComTQA contains more complex table questions in real-world table images to assess the robustness of various models.
As shown in Fig.~\ref{fig:com_tqa2}, we showcase several representative examples, including multiple answers, mathematical calculation and logical inference, which are the question types lacking in previous benchmarks.
To this end, we hope that ComTQA could fill this gap and serve as a reasonable benchmark for community development.
\vspace{-1.0em}

\vspace{-1.5em}

\section{Annotation in TSR task}
\label{appendix:annotation}
\begin{figure}[!h]
	\centering
	\includegraphics[width=1.0\textwidth]{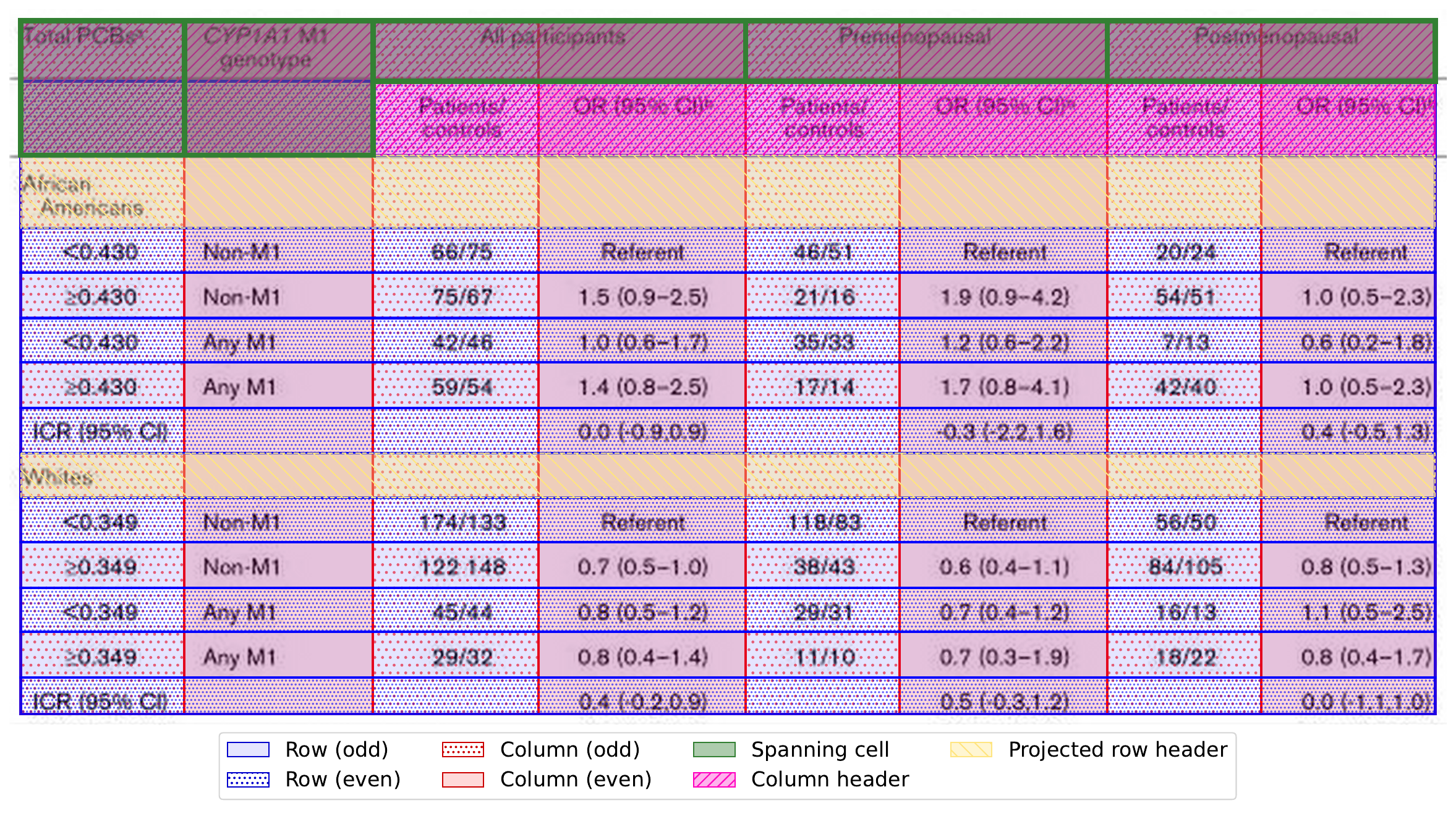} 
	\vspace{-1.0em}
	\caption{The illustration of an example table with dilated bounding box annotations for different object classes for modeling table structure recognition.}
	\label{fig:annotation}
\end{figure}

We illustrate the object classes utilized in TSR and TQ tasks as shown in Fig.~\ref{fig:annotation}.
A table generally is composed of five basic elements, i.e., column, row, spanning cell, column header and projected row header. "Row" denotes the rectangular boxes of each row's content in the table, while "Column" denotes the rectangular boxes of each column's content. The area where each row and each column intersect represents the table cell. Besides these both most common table elements, "Column header" refers to the area in the table that contains the data type or content for each column, usually occupying multiple rows at the top of the table. “Projected row header”, as a special row, represents the area that contains a single non-blank cell in a row. "Spanning cell" refers to a cell in a table that spans multiple rows or columns. According to these definitions, these objects have implicit relationship and construct a table’s hierarchical structure through physically overlapped rectangle boxes

\section{Broader Impact}
\label{Broader Impact}
Our proposed model targets to unify multiple visual form comprehension tasks.
This technology could help more people with visual impairments access tabular data through cooperating with improved screen readers and other assistive technologies.
Moreover, automating table understanding technology could reduce the need for time-consuming manual data entry and correction, freeing up human resources for more complex and creative tasks.
To be honest, this technology also brings some negative societal impacts. 
As more table data is extracted and processed with automatic visual table understanding, there is a heightened risk of sensitive information being mishandled or exposed. It is crucial to ensure robust data privacy measures.

\section{More Qualitative Results}
\label{More_Qualitative}
\begin{figure}[!h]
     \centering
	\includegraphics[width=0.95\textwidth]{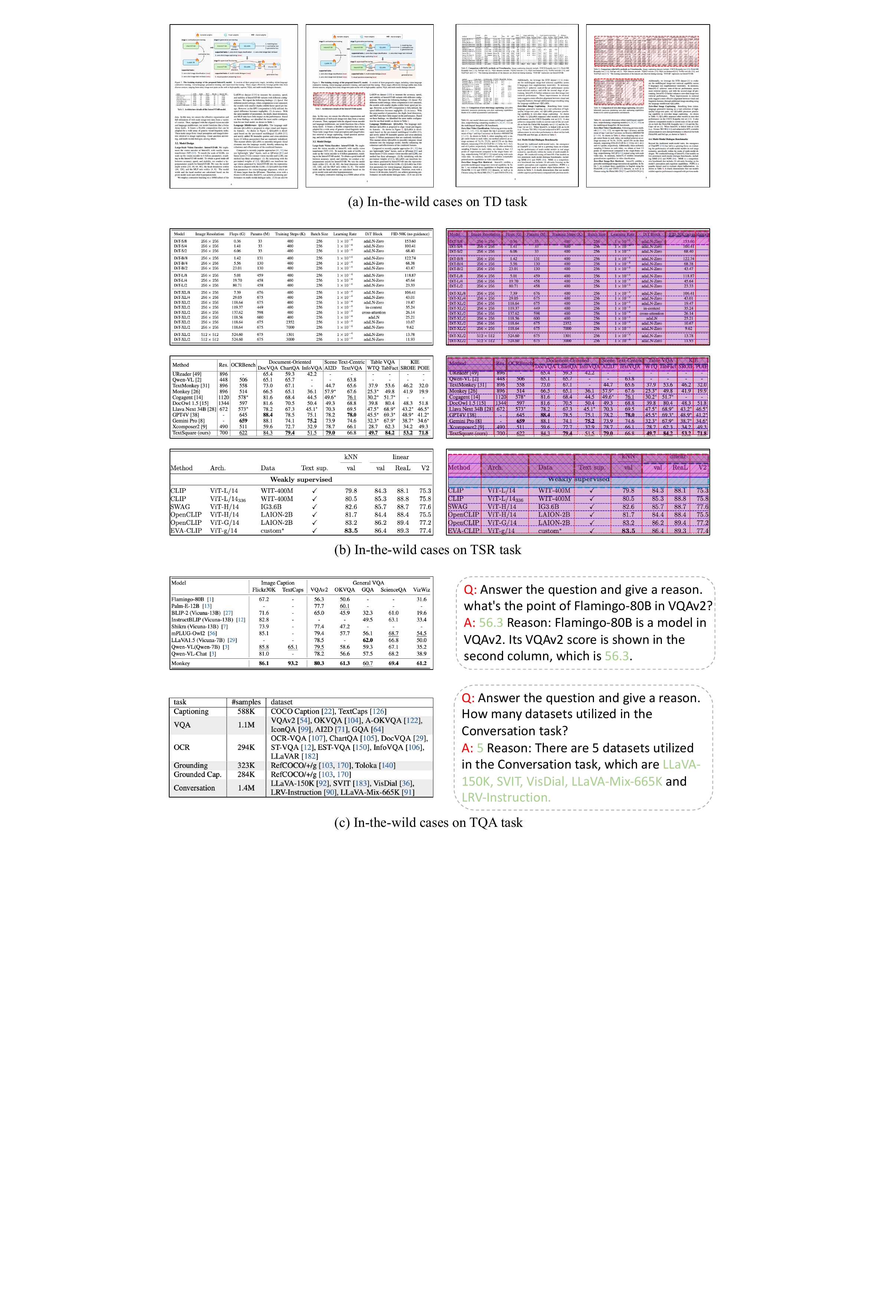} 
	\vspace{-0.5em}
	\caption{Qualitative results of TabPedia on in-the-wild cases. TabPedia achieves impressive performance in these unseen images, which validates its robustness and generalization. Zoom in for best view.}
	\label{fig:supp2}
\end{figure}

\textbf{Results on in-the-wild cases.} For better investigating the generalization of our proposed TabPedia, we randomly select some document images from a \href{https://arxiv.org/}{document website} and illustrate the generation results in Fig.~\ref{fig:supp2}. For perception and comprehension tasks, TabPedia generates accurate and reasonable responses in TD, TSR and TQA tasks, which sufficiently proves the robustness of our method for visual table understanding.

\begin{figure}[!h]
     \centering
	\includegraphics[width=0.75\textwidth]{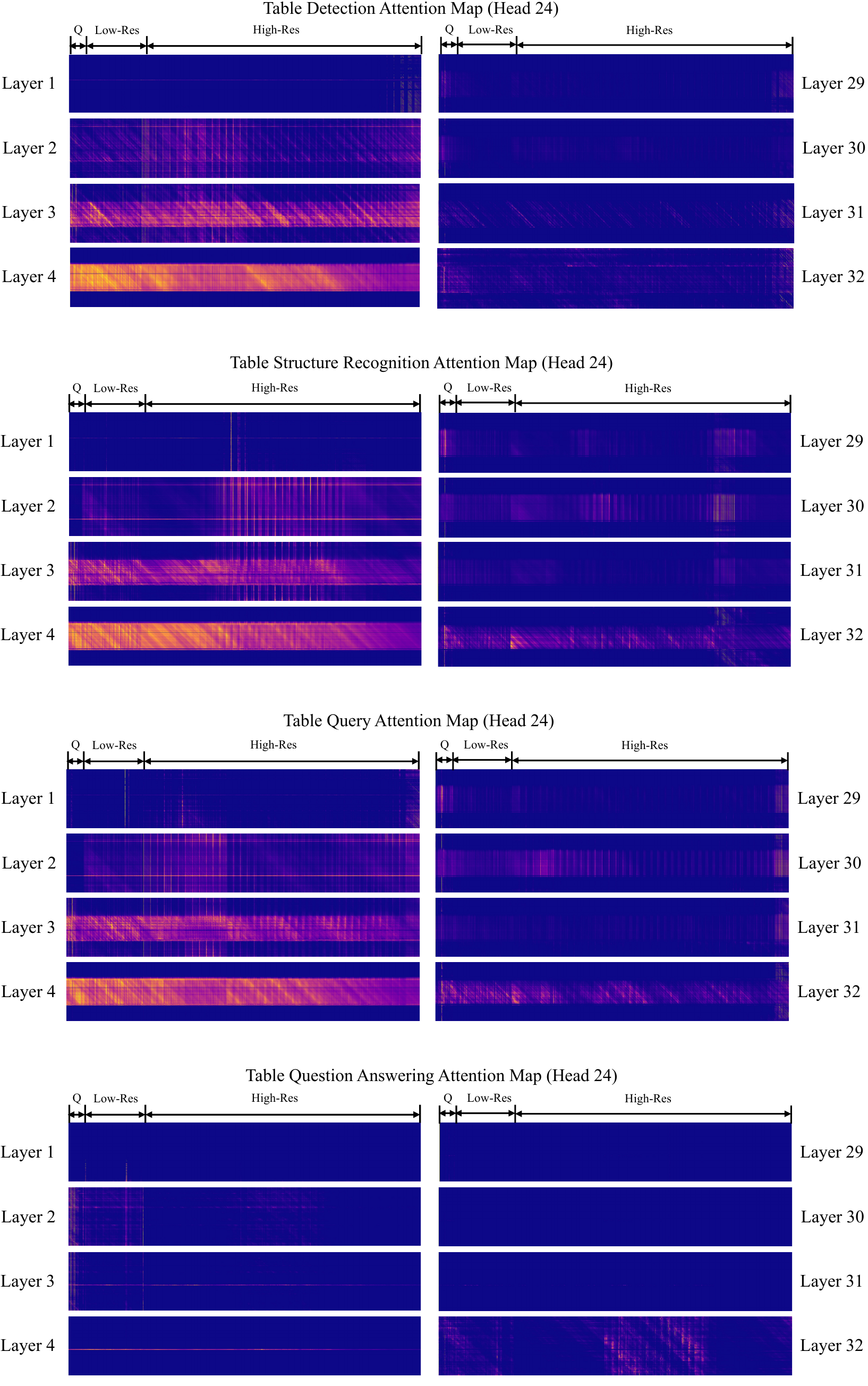} 
	\caption{Visualization of attention maps between meditative tokens and the sequence of instruction and visual tokens. ``Q'', ``Low-Res'' and ``High-Res'' denote the instruction tokens, global visual tokens and local visual tokens, respectively. Y-axis denotes the meditative tokens. Zoom in for best view.}
	\label{fig:atten_map}
\end{figure}

\textbf{Attention map of meditative tokens.} In order to analyze the information extraction of meditative tokens for different tasks, we visualized the attention maps of meditative tokens for input instructions with different granularity of visual feature tokens, as shown in Fig.~\ref{fig:atten_map}. For each task, we select the shallow and deep four-layer attention maps in the LLM for visualization, respectively. 
The y-axis represents the meditative tokens, while the x-axis represents the sequence of instruction tokens and different granular visual tokens.
For perceptive tasks, meditative tokens are densely attentive to most of the input information in the shallow layers, while they showcase diverse attention regions in the deeper layers.This phenomenon illustrates that meditative tokens could adaptively capture task-related information with respect to diverse tasks.
For the comprehension task (TQA), meditative tokens show a different attention pattern from perception tasks, which maintain sparse attention with input tokens in the shallow layers. 
These results validate that our proposed meditative tokens adaptively enable different regions of visual tokens and understand the intention of specific task questions.

\textbf{Visualization of TabPedia's responses.} As shown in Tab.~\ref{Tab:quality results},we introducing meditative tokens can bring promising performance across VTU tasks. We compare in detail the differences in the generated results before and after the introduction of Meditative in different VTU tasks. It is observed that introducing meditative tokens mainly improves the quality of long-form responses. Also for the perception tasks including TD and TSR, introducing meditative tokens can alleviate the meaningless or repetitive word generation. For the comprehension task, TQA, introducing meditative tokens can generate more elaborated and reasonable response. As suggested, we showcase several samples for better understanding.

\begin{table}[!h]
    \footnotesize
    \tabcolsep=6pt
    \tabcaption{Qualitative results of TabPedia's responses.}
    \label{Tab:quality results}
    \renewcommand\arraystretch{1.04}
    \resizebox{\linewidth}{!}{
        \begin{tabular}{c|p{3cm}<{\centering}|p{5cm}<{\centering}|p{6cm}<{\centering}|p{8cm}<{\centering}}	
            \toprule
            \textbf{Image} & \textbf{Question} & \textbf{GT}  & \textbf{TabPedia (w/o Meditative Tokens)}  & \textbf{TabPedia}\\        \midrule
                 \begin{minipage}[b]{0.3\columnwidth}
		\raisebox{-.5\height}{\includegraphics[width=\linewidth]{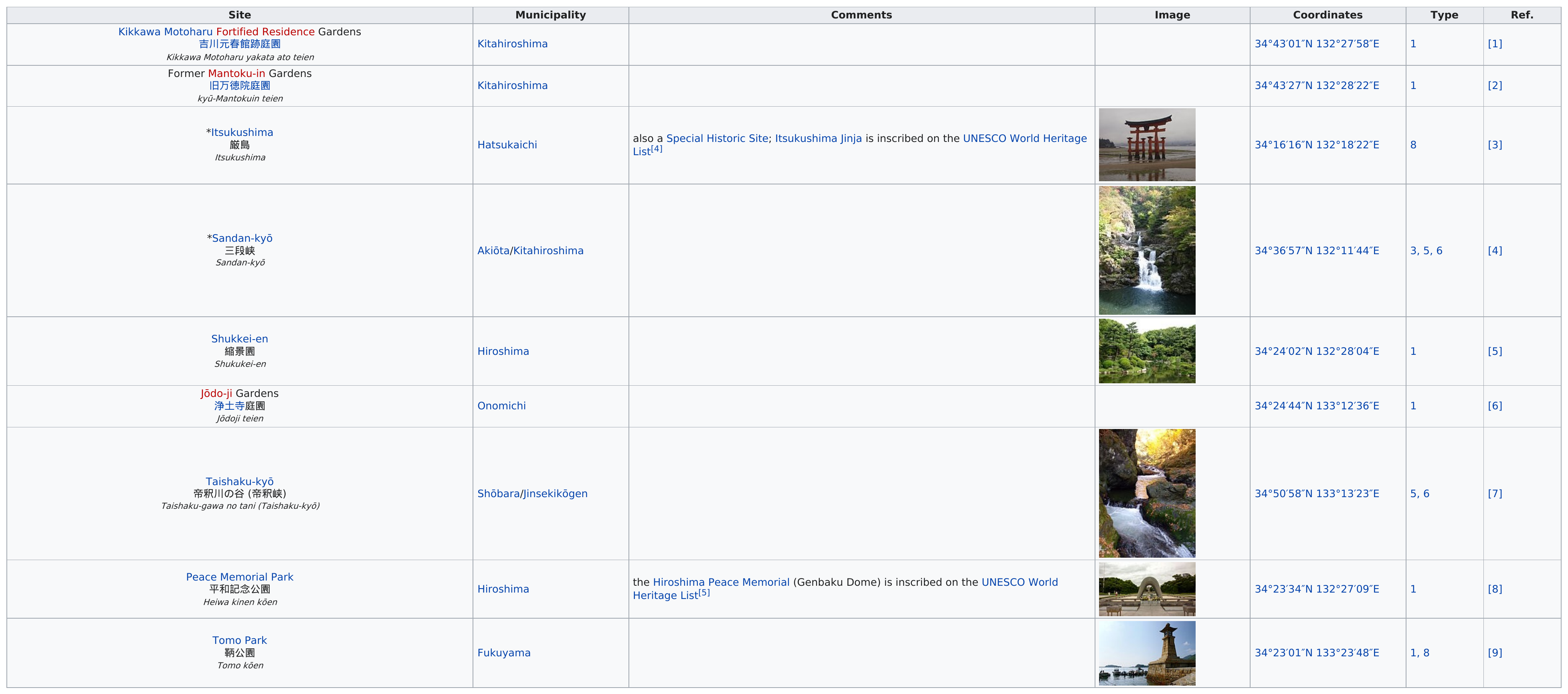}}
	\end{minipage} & what is the last site's municipality? & ``Fukuyama'' & ``Fukuoka \textbackslash n Reason: The last site is Tachibana, and its municipality is Fukuoka.'' & ``Fukuyama \textbackslash nReason: It is shown in the last row of the table that the last site's municipality is Fukuyama. So the answer is Fukuyama.'' \\
                \midrule
                 \begin{minipage}[b]{0.3\columnwidth}
		\raisebox{-.5\height}{\includegraphics[width=\linewidth]{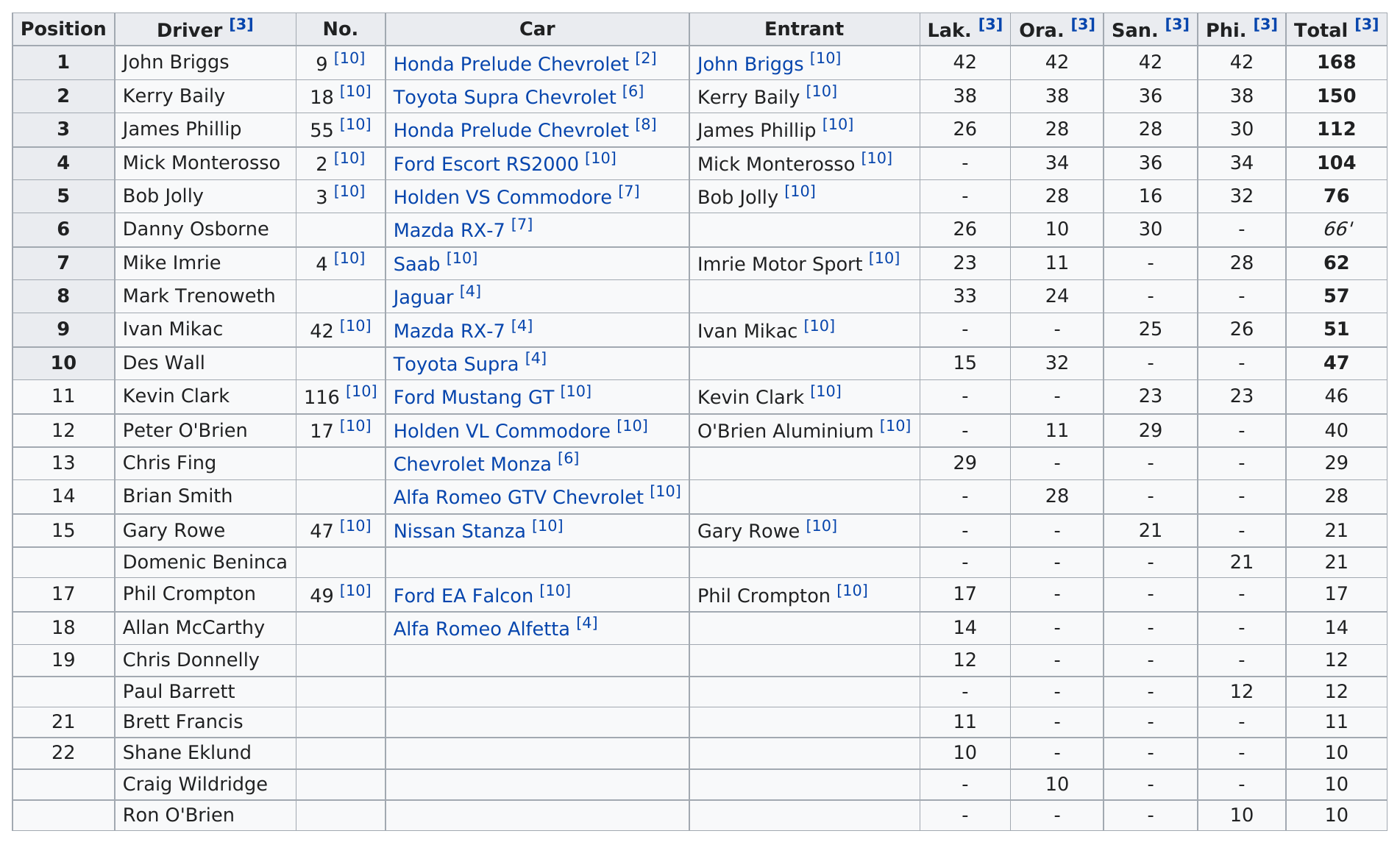}}
	\end{minipage}& john briggs and james phillip both drove which type of car? & ``Honda Prelude Chevrolet'' & ``Mazda RX-7 \textbackslash n Reason: From the image, we can see that both John Briggs and James Phillipps drove Mazda RX-7.'' & ``Honda Prelude Chevrolet \textbackslash n Reason: From the image, we can see that both John Briggs and James Phillip drove the Honda Prelude Chevrolet. John Briggs drove the No.9 Honda Prelude Chevrolet, while James Phillip drove the No.10 Honda Prelude Chevrolet. So they both drove the same type of car, which is the Honda Prelude Chevrolet.'' \\
                \midrule
                \begin{minipage}[b]{0.3\columnwidth}
		\raisebox{-.5\height}{\includegraphics[width=\linewidth]{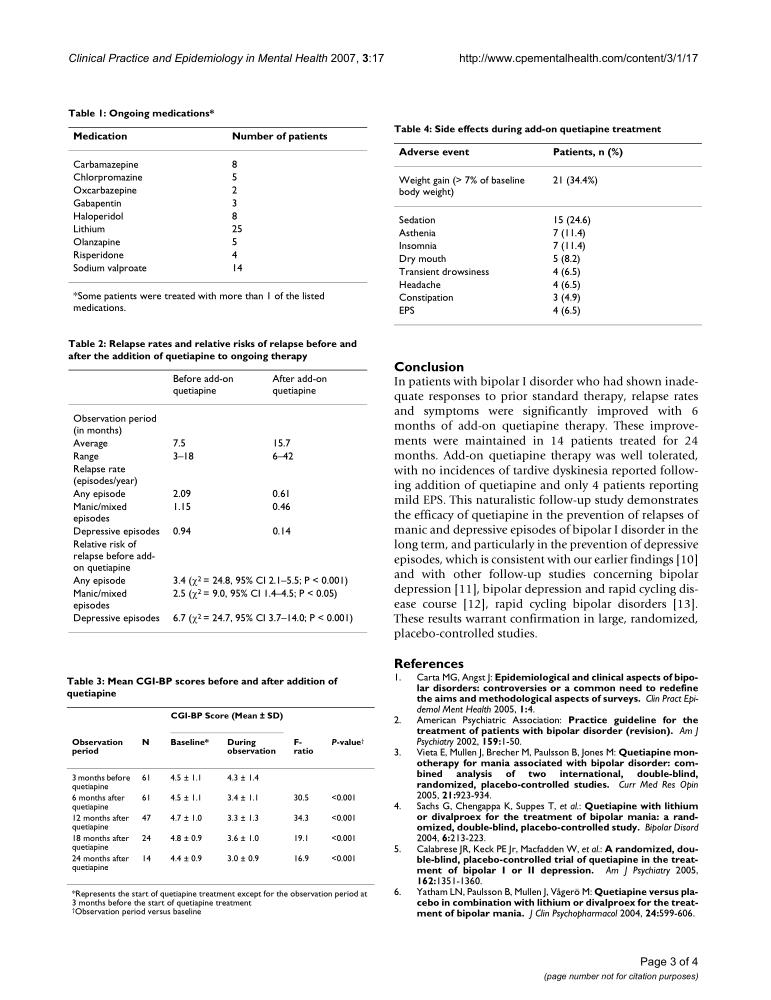}}
	\end{minipage}& Output all the table element's locations in the photo. & ``Here is a list of all the locations of table element in the picture:\textbackslash n [0.095,0.139,0.424,0.279]\textbackslash n [0.095,0.375,0.458,0.620]\textbackslash n [0.092,0.704,0.472,0.862]\textbackslash n [0.518,0.155,0.807,0.321]'' & ``Here is a list of all the locations of table element in the picture:\textbackslash n [0.095,0.139,0.426,0.329]\textbackslash n [0.095,0.409,0.447,0.669]\textbackslash n [0.095,0.699,0.459,0.859]'' & ``Here is a list of all the locations of table element in the picture:\textbackslash n [0.096,0.140,0.422,0.281]\textbackslash n [0.095,0.378,0.456,0.617]\textbackslash n [0.094,0.707,0.474,0.862]\textbackslash n [0.518,0.156,0.809,0.324]'' \\
            \bottomrule
        \end{tabular}
}
\end{table}

\end{document}